\documentclass[11pt]{article}

\usepackage[hyperref]{emnlp2021}
\usepackage{pgfplots}

\newcommand{\tabincell}[2]{\begin{tabular}{@{}#1@{}}#2\end{tabular}}

\usepackage[normalem]{ulem}
\usepackage{enumitem}
\usepackage{times}
\usepackage{latexsym}
\usepackage{amssymb}
\usepackage{amsmath} 
\usepackage{amstext}
\usepackage{booktabs}
\usepackage{enumerate}
\usepackage{graphicx}
\usepackage{subfigure}
\usepackage{xspace}
\usepackage{float}
\usepackage{bbm}
\usepackage{bm}
\usepackage{multirow}
\usepackage{booktabs}
\usepackage{color}
\usepackage{framed}
\usepackage{stfloats}
\usepackage{iitem}
\usepackage[T1]{fontenc}
\definecolor{shadecolor}{RGB}{180,180,180}
\usepackage{url}
\newcommand{\paratitle}[1]{\vspace{1.5ex}\noindent\textbf{#1}}
\newcommand{\ie}{\emph{i.e.,}\xspace}

\newcommand{\eg}{\emph{e.g.,}\xspace}

\newcommand{\ignore}[1]{}

\title{RocketQAv2: A Joint Training Method for Dense Passage Retrieval \\ and Passage Re-ranking}

\author{ \textbf{
Ruiyang Ren\textsuperscript{1,3}\thanks{\llap{}\:\:\:Equal contribution.\:\:The work was done when Ruiyang Ren was doing internship at Baidu.} ,
Yingqi Qu\textsuperscript{2}\footnotemark[1] , 
Jing Liu\textsuperscript{2},
Wayne Xin Zhao\textsuperscript{3,4}\thanks{\llap{}\:\:\:Corresponding authors. } , 
Qiaoqiao She\textsuperscript{2}
} \\
\textbf{
Hua Wu\textsuperscript{2}\footnotemark[2] ,
Haifeng Wang\textsuperscript{2} and Ji-Rong Wen\textsuperscript{1,3,4}
}\\
	\textsuperscript{1}School of Information, Renmin University of China;  
	\textsuperscript{2}Baidu Inc. \\
	\textsuperscript{3}Beijing Key Laboratory of Big Data Management and Analysis Methods\\
	\textsuperscript{4}Gaoling School of Artificial Intelligence, Renmin University of China\\ 
	\{reyon.ren, jrwen\}@ruc.edu.cn, batmanfly@gmail.com\\
	\{quyingqi, liujing46, sheqiaoqiao, wu\_hua, wanghaifeng\}@baidu.com
}

\begin{document}
\maketitle
\begin{abstract}
In various natural language processing tasks, passage retrieval and passage re-ranking are two key procedures in finding and ranking relevant information.
Since both the two procedures contribute to the final performance, it is important to jointly optimize them in order to achieve mutual improvement.
In this paper, we propose a novel joint training approach for dense passage retrieval and passage re-ranking.
A major contribution is that we introduce the dynamic listwise distillation, where we design a unified listwise training approach for both the retriever and the re-ranker. 
During the dynamic distillation, the retriever and the re-ranker can be adaptively improved according to each other's relevance information. 
We also propose a hybrid data augmentation strategy to construct diverse training instances for listwise training approach. 
Extensive experiments show the effectiveness of our approach on both MSMARCO and Natural Questions datasets. Our code is available at ~\url{https://github.com/PaddlePaddle/RocketQA}.



\end{abstract}

\section{Introduction}
\label{section:intro}
Recently, dense passage retrieval has become an important approach in the task of \emph{passage retrieval}~\citep{zhao2022dense, dpr2020} to identify relevant contents from a large corpus. The underlying idea is to represent both queries and passages as low-dimensional vectors (a.k.a., embeddings), so that the relevance can be measured via embedding similarity. Additionally, a subsequent procedure of \emph{passage re-ranking} is widely adopted to further improve the retrieval results by incorporating a re-ranker~\citep{rocketqa, MEBERT}. Such a two-stage procedure is particularly useful in a variety of natural language processing tasks, including question answering~\citep{mao2021reader,multihopqa}, dialogue system~\citep{conversation,henderson2017efficient} and entity linking~\citep{entitylinking,wu2019scalable}. 


Following a \emph{retrieve-then-rerank} way, the dense retriever in passage retrieval and the re-ranker in passage re-ranking jointly contribute to the final performance. 
Despite the fact that the two modules work as a pipeline during the inference stage, it has been found useful to train them in a correlated manner. 
For example, the retriever with a dual-encoder can be improved by distilling from the re-ranker with a more capable cross-encoder architecture~\cite{rocketqa, google2020augmentation}, and the re-ranker can be improved with training instances generated from the retriever~\cite{rocketqa, Huang2020embedding, gao2021rethink}. 
Therefore, there is increasing attention on correlating the training of the retriever and re-ranker in order to achieve mutual improvement~\cite{metzler2021rethinking, rocketqa, Huang2020embedding, google2020augmentation}. 
Typically, these attempts train the two modules in an alternative way: fixing one module and then optimizing another module.
It will be more ideal to mutually improve the two modules in a joint training approach. 

However, the two modules  are usually optimized in different ways, so that the joint learning cannot be trivially implemented. 
Specially, the retriever is usually trained by sampling a number of in-batch negatives to maximize the probabilities of positive passages and minimize the probabilities of the sampled negatives~\citep{ANCE,dpr2020}, where the model is learned by considering the entire list of positive and negatives (called \emph{listwise} approach\footnote{Instead of considering the total order as in learning to rank~\citep{cao2007learntorank}, we use ``listwise'' to indicate that relevance scores are derived based on a candidate list.}). As a comparison, the re-ranker is usually learned in a pointwise or pairwise manner~\citep{nogueira2019passage,multistage}, where the model is learned based on a single passage or a pair of passages. 
To address this issue, our idea is to unify the learning approach for both retriever and re-ranker.
 Specially, we adopt the listwise training approach for both retriever and re-ranker, where the relevance scores are computed according to a list of positive and negative passages. 
Besides, it is expected to include diverse and high-quality training instances for the listwise training approach, which can better represent the distribution of all the passages in the whole collection. Thus, it requires more effective data augmentation to construct the training instances for listwise training.

To this end, we present a novel joint training approach for dense passage retrieval and passage re-ranking~(called \textbf{RocketQAv2}). 
The major contribution of our approach is the novel \emph{dynamic listwise distillation} mechanism for jointly training the retriever and the re-ranker. Based on a unified listwise training approach, we can readily transfer relevance information between the two modules.
Unlike previous distillation methods that usually froze one module, our approach enables the two modules to adaptively learn relevance information from each other, which is the key to mutual improvement in joint training. Furthermore, we design a hybrid data augmentation strategy to generate diverse training instances for listwise training approach. 

The contributions of this paper can be summarized as follows:
\begin{itemize}
    \item We propose a novel approach that jointly trains the dense passage retriever and passage re-ranker. It is the first time that joint training has been implemented for the two modules.
    \item We make two major technical contributions by introducing dynamic listwise distillation and hybrid data augmentation to support the proposed joint learning approach.
    \item Extensive experiments show the effectiveness of our proposed approach on both MSMARCO and Natural Questions datasets. 
\end{itemize}
\section{Related Work}

Recently, dense passage retrieval has demonstrated better performance than traditional sparse
retrieval methods (e.g., TF-IDF and BM25) on the task of passage retrieval. Existing approaches of learning dense passage retriever can be divided into two categories: (1)
self-supervised pre-training for retrieval~\cite{pretrain2020iclr, latent2019acl, REALM}  and (2)
fine-tuning pre-trained language models (PLMs) on labeled data~\cite{negative2020google, dpr2020, ANCE, MEBERT, rocketqa} . Our work follows the second class of approaches, which show better performance with less
cost. There are two important tricks to train an effective dense retriever: (1) incorporating hard negatives during training~\citep{dpr2020, ANCE, rocketqa} and (2) distilling the knowledge from cross-encoder-based reranker into dual-encoder-based retriever~\citep{izacard2020distilling, yang2020retriever, rocketqa, pair}. 
Based on the retrieved passages from a retriever, PLM-based rerankers with the cross-encoder architecture have recently been applied on passage re-ranking to improve the retrieval results~\citep{qiao2019understanding, nogueira2019passage, wang2019multi, yan2019idst}, and yield substantial improvements over the traditional methods. 


Apart from separately considering the above two tasks, it has been proved that passage retrieval and passage re-ranking are actually highly related and dependent~\citep{Huang2020embedding,Gao2020ModularizedTR,colbert2020sigir}. The retriever needs to capture the relevance knowledge from the re-ranker, and the re-ranker should be specially optimized according to the preceding results of the retriever. Some efforts studied the possibility of leveraging the dependency of retriever and re-ranker, and try to enhance the connection between them in an alternative way~\cite{rocketqa,google2020augmentation,Huang2020embedding}. Furthermore, several studies attempted to jointly train the retriever and the reader for Open-domain Question Answering~\cite{REALM, sachan2021end, dpr2020}.
Different from the prior studies, our method is a joint learning architecture of the dense passage retriever and the re-ranker. 

\section{Methodology}
In this section, we describe a novel joint training approach for dense passage retrieval and passage re-ranking~(called \textbf{RocketQAv2})

\subsection{Overview}
\label{section:overview}
In this work, we consider two tasks including dense passage retrieval and passage re-ranking, which are described as follows.

Given a query $q$, the aim of \emph{dense passage retrieval} is to retrieve $k$ most relevant passages from a large collection of $M$ text passages. The dual-encoder~(DE) architecture is widely adopted by prior works~\citep{dpr2020, MEBERT, rocketqa}, where two separate dense encoders $E_P(\cdot)$ and $E_Q(\cdot)$ are used to map passages and queries to $d$-dimensional real-valued vectors (a.k.a., embeddings) separately, and then an index of all passage embeddings is built for efficient retrieval. The similarity between the query $q$ and the passage $p$ is defined using the dot product:
\begin{equation}
\label{equation:sim_definition}
s_\text{de}(q, p)=E_Q(q)^{\top} \cdot E_P(p).
\end{equation}
\ignore{
Let $p^+$ denotes positive passage and $p^-$ denotes negative passage, the above similarity are usually learned by optimizing the negative log likelihood of the positive passage:
\begin{equation}
\label{equation:L_Q}
\small
L = -\frac{1}{N} \sum_{\langle q, p+\rangle} \log \frac{e^{s_\text{de}(q, p^+)}}{e^{s_\text{de}(q, p^+)} + \sum_{p^-} e^{s_\text{de}(q, p^-)}}.
\end{equation}
Prior works often adopt in-batch negative sampling to obtain amount of in-batch negatives~(weak negatives) to boost the performance of dual-encoder~\citep{dpr2020,MEBERT}. 
}

Given a list of candidate passages retrieved by a passage retriever, the aim of \emph{passage re-ranking} is to further improve the retrieval results with a re-ranker, which estimates a relevance score $s(q,p)$ measuring the relevance level of a candidate passage $p$ to a query $q$. 
Among the implementations of the re-ranker, a cross-encoder~(CE) based on PLMs usually achieves superior performance~\citep{nogueira2019passage, qiao2019understanding}, which can better capture the semantic interactions between the passage and the query, but requires more computational efforts than the dual-encoder. 
To compute the relevance score $s_\text{ce}(q,p)$, a special token \textsc{[SEP]} is inserted between $q$ and $p$, and then the representation at the \textsc{[CLS]} token from the cross-encoder is fed into a learned linear function. 

Usually, the passage retriever and the passage re-ranker are learned in either a separate or alternative way (\ie fixing one and then training the other). 
To achieve the joint training, we introduce  \emph{dynamic listwise distillation}~(Section~\ref{section:distill}), which can adaptively improve both components in a joint optimization process.  To support the listwise training, we further propose \emph{hybrid data augmentation} (Section~\ref{section:distill})  for generating diverse and high-quality training instances.
Based on the two major contributions, we present the learning procedure in Section~\ref{section:procedure} and related discussion in Section~\ref{section:discussion}.  

\begin{figure}
	\centering 
	\includegraphics[width=0.49\textwidth]{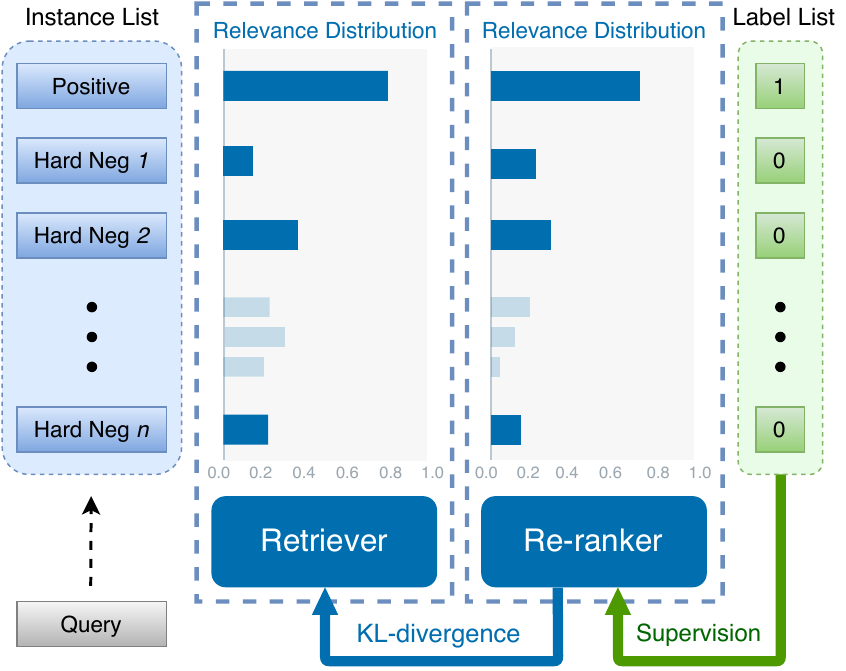}
	\caption{The illustration of dynamic listwise distillation in our approach. }
	\label{fig:model} 
\end{figure}

\ignore{Previous studies trained the retriever and the re-ranker individually. However, both retriever and re-ranker are actually highly related and dependent, the score distributions from retriever and re-ranker need to be referred to by each other for optimal performance. Therefore, we propose the joint training architecture for dense passage retrieval and passage re-ranking, including two technical contributions: Dynamic distillation couples the two modules by the score distribution, hybrid data augmentation method enables the joint training architecture to touch hard negatives from different sampling approach as a guarantee for dynamic distillation. Figure~\ref{fig:model} presents the overall architecture. }

\subsection{Dynamic Listwise Distillation}
\label{section:distill}

Since the re-ranker adopts the more capable cross-encoder architecture, it has become a common strategy to distill the knowledge from re-ranker into the retriever.  
However, in prior studies~\cite{dpr2020, ANCE, rocketqa}, the retriever and re-ranker are usually learned in different ways, and the parameters of the re-ranker are \emph{frozen}, which cannot jointly optimize the two components for mutual improvement. 
Considering this issue, we design a unified listwise training approach to learn both the retriever and the re-ranker, and dynamically update both the parameters of the re-ranker and the retriever during distillation. In this way, the two components can adaptively improve each other. We call this approach as \emph{dynamic listwise distillation}. Next, we will describe the details of \emph{dynamic listwise distillation}. 





Formally, given a query $q$ in a query set $\mathcal{Q}$ and the corresponding list of candidate passages~(instance list) $\mathcal{P}_q = \{p_{q,i}\}_{1 \leq i \leq m}$ related to query $q$, 
we can obtain the relevance scores $S_\text{de}(q) = \{s_\text{de}(q, p)\}_{p\in \mathcal{P}_q}$ and $S_\text{ce}(q) = \{s_\text{ce}(q, p)\}_{p\in \mathcal{P}_q}$ of a query $q$ and passages in $\mathcal{P}_q$ from the dual-encoder-based retriever and the cross-encoder-based re-ranker, respectively. 
Then, we normalize them in a listwise way to obtain the corresponding relevance distributions over candidate passages: 
\begin{eqnarray}
\tilde s_\text{de}(q, p) &=& \frac{e^{s_\text{de}(q, p)}}{\sum_{p' \in \mathcal{P}_q} e^{s_\text{de}(q, p')}},\label{equation:kl_divergence}\\
\tilde s_\text{ce}(q, p) &=& \frac{e^{s_\text{ce}(q, p)}}{\sum_{p' \in \mathcal{P}_q} e^{s_\text{ce}(q, p')}}\label{equation:kl_divergence}.
\end{eqnarray}

The main idea is to 
adaptively reduce the  difference between the two distributions from the retriever and the re-ranker so as to mutually improve each other.   
 To achieve the adaptively mutual improvement, we minimize the KL-divergence between the two relevance distributions $\{\tilde s_\text{de}(q, p)\}$ and $\{\tilde s_\text{ce}(q, p)\}$ from the two modules:
\begin{equation}
\label{equation:kl_divergence}
\mathcal{L}_{\text{KL}} = \sum_{q \in \mathcal{Q}, p \in \mathcal{P}_q}
\tilde s_\text{de}(q, p) \cdot {\log \frac{\tilde s_\text{de}(q,p)}{\tilde s_\text{ce}(q, p)}}.
\end{equation}

Additionally, we provide ground-truth guidance for the joint training. Specifically, we also adopt a cross-entropy loss for the re-ranker based on passages in $\mathcal{P}_q$ with supervised information:
\begin{equation}
\label{equation:sup}
\small
\mathcal{L}_\text{sup} = -\frac{1}{N} \sum_{q \in \mathcal{Q}, p^+} \log \frac{e^{s_\text{ce}(q, p^+)}}{e^{s_\text{ce}(q, p^+)} + \sum_{p^-} e^{s_\text{ce}(q, p^-)}},
\end{equation}
where $N$ is the number of training instances, and $p^+$ and $p^-$ denote the positive passage and negative passage in $\mathcal{P}_q$, respectively.
We combine the KL-divergence loss and the supervised cross-entropy loss defined in Eq.~(\ref{equation:kl_divergence}) and Eq.~(\ref{equation:sup}) to obtain the final loss function:
\begin{equation}
\label{equation:tot_loss}
\mathcal{L}_{\text{final}} = \mathcal{L}_{\text{KL}} + \mathcal{L}_{\text{sup}}.
\end{equation}

Figure~\ref{fig:model} presents the illustration of the dynamic listwise distillation. The re-ranker is optimized with labeled lists (Eq.~(\ref{equation:sup})), and it produces relevance distributions to train the retriever (Eq.~(\ref{equation:kl_divergence})).  
Unlike RocketQA that conducts hard pseudo labeled data~\citep{rocketqa}, we utilize soft labels (\ie estimated relevance distributions) for relevance distillation. Besides, we dynamically update the parameters of the re-ranker in order to adaptively synchronize the two modules for mutual improvement.
To discriminate from the previous static distillation based on pseudo labels, we call our method as \emph{dynamic listwise distillation}.

\begin{figure}
	\centering 
	\includegraphics[width=0.49\textwidth]{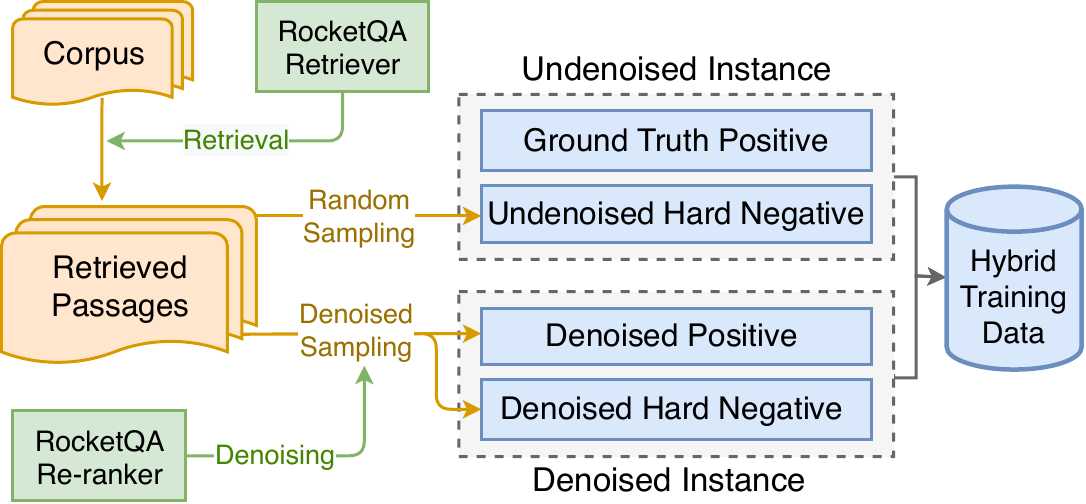}
	\caption{The illustration of hybrid data augmentation. }
	\label{fig:data} 
\end{figure}

\subsection{Hybrid Data Augmentation}
\label{section:negative-sampling}
To perform dynamic listwise distillation, we need to generate the candidate passage list $\mathcal{P}_q$ for query $q$. Since our approach relies on listwise training, we expect the candidate passage list includes diverse and high-quality candidate passages, which may better represent the distribution of all the passages in the whole collection. Prior works~\citep{ANCE, rocketqa, dpr2020} demonstrate that it is important to include hard negatives in the candidate passage list. Basically, ANCE~\citep{ANCE} and DRP~\citep{dpr2020} introduces the randomly sampled hard negatives, while RocketQA~\citep{rocketqa} incorporates denoised hard negatives. Inspired by prior works, we design a hybird data augmentation way to construct diverse training instances by incorporating both random sampling and denoised sampling. 


As shown in Figure~\ref{fig:data}, our proposed hybrid data augmentation includes both undenoised and denoised instances. First, we utilize the RocketQA retriever to retrieve top-$n$ passages from the corpus. 
For undenoised instances, we randomly sample the undenoised hard negatives from retrieved passages  and include ground-truth  positives. For denoised instances, we utilize the RocketQA re-ranker to remove the predicted negatives with low confidence scores.  We also include denoised positives that are predicted as positives by the RocketQA re-ranker with high confidence scores. 




\ignore{
The first type of instances including labeled passages as positives and a random subset of recalled passages by the existing retriever as hard negatives. Concretely, we use the step-$1$ dual-encoder in RocketQA~\cite{rocketqa} to recall amount of related passages~(\eg top-1000) of query $q$. We randomly sample $n$ passages from recalled passages as hard negatives.
Moreover, since top-retrieved passages that are likely to be false negatives, we use the denoised hard negatives and positives as the second type of instances similar to prior work~\citep{rocketqa}. We utilize the step-$1$ cross-encoder to filter positives and hard negatives, note that this part of positives are not leveraged in RocketQA. Finally, we use one positive passage and $n$ hard negative passages in the sampled passage list $\mathcal{P}_q$ of the query $q$.
}

Compared with previous methods, our data augmentation method utilizes more ways (undenoised or denoised) to generate both positives and negatives to improve the diversity of instances list $\mathcal{P}_q$. Specially, we mainly focus on including hard negatives. This is particularly important to dynamic listwise distillation, since weak negatives are easy to be identified, which cannot increase additional gain for both modules.




\subsection{Training Procedure}
\label{section:procedure}

In this section, we present the training procedure of our approach.

\begin{figure}
	\centering 
	\includegraphics[width=0.45\textwidth]{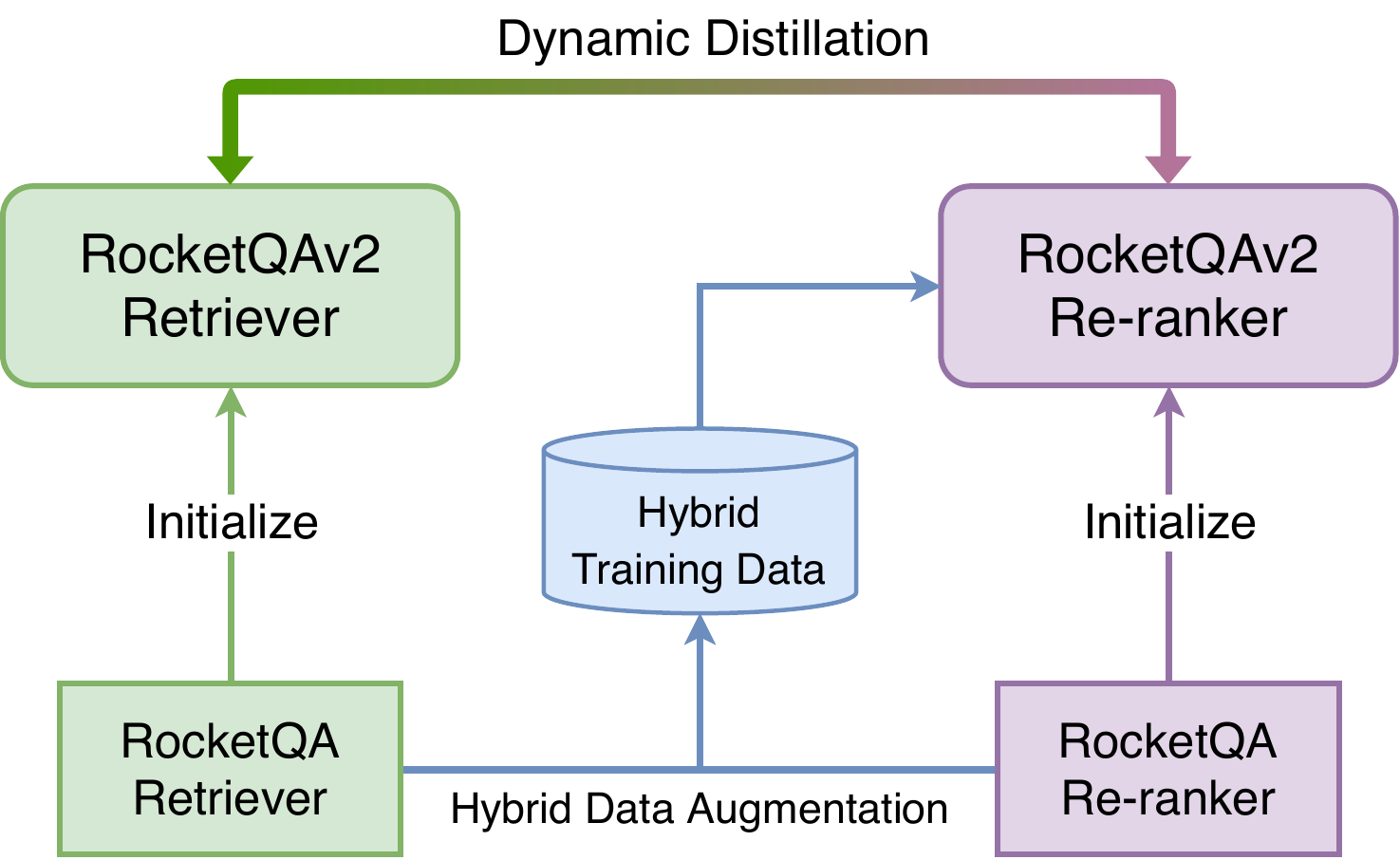}
	\caption{The overall joint training architecture of RocketQAv2.}
	\label{fig:pipeline} 
\end{figure}

Figure~\ref{fig:pipeline} presents the illustration of the training procedure for our approach.
We first initialize the retriever and re-ranker with the learned dual-encoder and cross-encoder of RocketQA~\footnote{Note that in this paper, RocketQA retriever is the model in the first step of RocketQA and the RocketQA re-ranker is the model in the second step of RocketQA. The two models can also be replaced with other trained retriever and re-ranker. We found that using the trained model to initialize retriever and reranker can help achieve slightly better results. This is due to the fact that the retriever and re-ranker have a mutual influence during training, the initialized retriever and re-ranker can facilitate the initial optimization stage. }.
Then, we utilize the retriever and re-ranker in RocketQA to generate the training data via hybrid data augmentation in Section~\ref{section:negative-sampling}. 
Finally, we preform dynamic listwise distillation to jointly optimize the retriever and re-ranker following Section~\ref{section:distill}.
During distillation, the retriever and re-ranker are mutually optimized according to the final retrieval performance. 
After the training stage, we can apply the retriever and re-ranker for inference in a pipeline.

\ignore{As shown in the Figure~\ref{fig:pipeline}, we organize the above two technical contribution into an effective training pipeline.

\begin{itemize}[noitemsep,topsep=0pt,parsep=0pt,partopsep=0pt,labelindent=0cm,leftmargin=0cm]
    \item \textbf{Step 1:} Construct both undenoised instance and denoised instance as hybrid training data by hybrid data augmentation.
    \item \textbf{Step 2:} Perform joint training of the retriever and the re-ranker with dynamic distillation. The warm-start models of joint training are RocketQA retriever and re-ranker. In dynamic distillation, the supervised signal of the retriever comes entirely from the re-ranker, since the poor ability of the warm start re-ranker will hinder the learning of retriever at beginning. So that, the joint training needs to start based on two modules that already have certain ability.
\end{itemize}
}


\subsection{Discussion}
\label{section:discussion}
In this section, we discuss the comparison with RocketQA. 

This work presents an extended contribution to RocketQA~\cite{rocketqa}, called RocketQAv2.
As  seen from above, RocketQAv2 reuse the network architecture and important training tricks in RocketQA. 
A significant improvement  is that RocketQAv2 incorporates a joint training approach for both the retriever and the re-ranker via dynamic listwise distillation.
For dynamic listwise distillation, RocketQAv2 designs a unified listwise training approach, and utilizes soft relevance labels for  mutual  improvement.
Such a distillation mechanism is able to simplify the training process,  and also provides the possibility for end-to-end training the entire dense retrieval architecture.

\begin{table*}[]
    \centering
    
    \begin{tabular}{c|c|c|c|c}
    \toprule
     \textbf{Dataset}   &  \textbf{\#query in train} & \textbf{\#query in dev} & \textbf{\#query in test} & \textbf{\#passage} \\
     \midrule
        MSMARCO & 502,939 &6,980 & 6.837 & 8,841,823 \\
        Natural Questions & 58,812 & 6,515 & 3,610 & 21,015,324 \\
    \bottomrule
    \end{tabular}
    \caption{The detailed statistics of MSMARCO and Natural Questions. }
    \label{tab:dataset}
\end{table*}

\section{Experiments}
In this section, we first describe the experimental settings, then report the main experimental results, ablation study, and detailed analysis.

\subsection{Experimental Setup}

\paratitle{Datasets}\quad
We adopt two public datasets on dense passage retrieval and passage re-ranking, including MSMARCO~\cite{msmarco} and Natural Questions~\cite{nq}. Table~\ref{tab:dataset} lists the statistics of the datasets. 
\emph{MSMARCO} was originally designed for multiple passage machine reading comprehension, and its queries were sampled from Bing search logs. Based on the queries and passages in MSMARCO Question Answering, MSMARCO Passage Ranking for passage retrieval and ranking was created.
\emph{Natural Questions (NQ)} was originally introduced for open-domain QA. This corpus consists of real queries from the Google search engine along with their long and short answer annotations from the top-ranked Wikipedia pages. DPR~\cite{dpr2020} selected the queries that had short answers and processed all the Wikipedia articles as the collection of passages. In our experiments, we reuse the NQ version created by DPR.

\paratitle{Evaluation Metrics}\quad Following previous work, we adopt Mean Reciprocal Rank (MRR) and Recall at top $k$ ranks (Recall@$k$) to evaluate the performance of passage retrieval. MRR calculates the averaged reciprocal of the rank at which the first positive passage is retrieved. Recall@$k$ calculates the proportion of questions to which the top $k$ retrieved passages contain positives.

\paratitle{Model Specifications}\quad
Our retriever and re-ranker largely follow ERNIE-2.0 base \citep{ernie20aaai}, which is a BERT-like~\citep{bert2019naacl} model with 12-layer transformers and introduces a continual pre-training framework on multiple pre-trained tasks. As described in previous section, 
the retriever is initialized with the parameters of the dual-encoder in the first step of RocketQA, and the re-ranker is initialized with the parameters of the cross-encoder in the second step of RocketQA. 

\paratitle{Implementation Details}\quad
We conduct experiments with the deep learning framework PaddlePaddle~\citep{ma2019paddlepaddle} on up to 32 NVIDIA Tesla V100 GPUs (with 32G RAM). 
For both two datasets, we used the Adam optimizer~\citep{adam} with a learning rate of 1e-5. 
The model is trained up to 3 epochs with a batch size of 96. The dropout rates are set to 0.1 on the cross-encoder. 
The ratio of the positive to the hard negative is set to 1:127 on MSMARCO and 1:31 on NQ.

\begin{table*}[htbp]
    \small
    \centering
    \begin{tabular}{llcccccc}
    \toprule
        \multirow{2}*{\textbf{Methods}} & \multirow{2}*{\textbf{PLM}} &  \multicolumn{3}{c}{\textbf{MSMARCO Dev}} & \multicolumn{3}{c}{\textbf{Natural Questions Test}} \\
               & & MRR@10 & R@50 & R@1000 & R@5 &  R@20 &  R@100\\
    \midrule
        BM25 (anserini)~\cite{bm25} & - & 18.7 & 59.2 & 85.7 & - & 59.1 & 73.7 \\
        \midrule
        doc2query~\cite{doc2query} & - & 21.5 & 64.4 & 89.1 & - & - & - \\
        DeepCT~\cite{deepct2019sigir} & - & 24.3 & 69.0 & 91.0 & - & - & - \\
        docTTTTTquery~\cite{doctttttquery} & - & 27.7 & 75.6 & 94.7 & - & - & - \\
        GAR~\cite{generation20augmented} & - & - & - & - & - & 74.4 & 85.3 \\
        UHD-BERT~\cite{uhdbert2021} & - & 29.6 & 77.7 & 96.1 & - & - & - \\        COIL~\cite{gao2021coil} & - & 35.5 & - & 96.3 & - & - & - \\
        \midrule
        DPR (single)~\cite{dpr2020} & BERT$_\text{base}$ & - & - & - & - & 78.4 & 85.4 \\
        DPR-E & ERNIE$_\text{base}$ & 32.5 & 82.2 & 97.3 &  68.4  & 80.7  & 87.3\\
        ANCE (single)~\cite{ANCE} & RoBERTa$_\text{base}$ & 33.0 & - & 95.9 & - & 81.9 & 87.5 \\
        TAS-Balanced~\cite{tas2021} & BERT$_\text{base}$ & 34.0 & - & 97.5 & - & - & - \\
        ME-BERT~\cite{MEBERT} & BERT$_\text{large}$ & 34.3 & - & - & - & - & - \\
        ColBERT~\cite{colbert2020sigir} & BERT$_\text{base}$ & 36.0 & 82.9 & 96.8 & - & - & - \\
        NPRINC~\cite{negative2020google} & BERT$_\text{base}$ & 31.1 & - & 97.7 & 73.3 & 82.8 & 88.4 \\
        ADORE+STAR~\cite{Optimizing2021sigir} & RoBERTa$_\text{base}$ & 34.7 & - & - & - & - & - \\
        RocketQA~\cite{rocketqa} & ERNIE$_\text{base}$ & 37.0 & 85.5 & 97.9 & 74.0 & 82.7 & 88.5 \\
        PAIR~\cite{pair} & ERNIE$_\text{base}$ & \underline{37.9} & \textbf{86.4} & \textbf{98.2} & \underline{74.9} & \underline{83.5} & \textbf{89.1}\\
        \midrule
        \textbf{RocketQAv2~(retriever)} & ERNIE$_\text{base}$ & \textbf{38.8} & \underline{86.2} & \underline{98.1} & \textbf{75.1}  & \textbf{83.7} & \underline{89.0} \\
    \bottomrule
    \end{tabular}
    \caption{Passage retrieval results on MSMARCO and Natural Questions datasets. PLM is the abbreviation of Pre-trained Language Model. We copy the results from original papers and we leave it blank if the original paper does not report the result. The best and second-best results are in bold and underlined fonts respectively.}
    \label{tab:main_results}
\end{table*}

\subsection{Results on Passage Retrieval}
In this part, we first describe the comparing baselines, then report the results on passage retrieval.

\subsubsection{Baselines}
To have comprehensive comparison, we choose as baselines the state-of-the-art approaches that consider both sparse and dense passage retrievers. 

The sparse retrievers include the traditional retriever BM25~\cite{bm25} and five traditional retrievers enhanced by neural networks, including doc2query~\cite{doc2query}, DeepCT~\cite{deepct2019sigir}, docTTTTTquery~\cite{doctttttquery}, GAR~\cite{generation20augmented}, UHD-BERT~\cite{uhdbert2021} and COIL~\cite{gao2021coil}. 
Both doc2query and docTTTTTquery employ neural query generation to expand documents. In contrast, GAR employs neural generation models to expand queries and UHD-BERT is empowered by extremely high dimensionality and controllable sparsity. Different from them, DeepCT and COIL utilizes BERT to learn the term weight or inverted list. 

The dense retrievers include DPR~\cite{dpr2020}, DPR-E, ANCE~\cite{ANCE}, ME-BERT~\cite{MEBERT}, NPRINC~\cite{negative2020google}, ColBERT~\cite{colbert2020sigir} RocketQA~\cite{rocketqa}, TAS-Balanced~\cite{tas2021}, ADORE+STAR~\citep{Optimizing2021sigir} and PAIR~\cite{pair}. 
DPR-E is our implementation of DPR using ERNIE~\cite{ernie20aaai} instead of BERT, which is to examine the effects of pre-trained language models.

\subsubsection{Results}
The results of different passage retrieval methods are presented in Table~\ref{tab:main_results}. It can be observed that:

(1) Among all methods, we can see the RocketQAv2 retriever and PAIR outperform other baselines by a large margin. PAIR is a contemporaneous work with RocketQAv2, which obtains improvement by pre-training on out-of-domain data. We observe that RocketQAv2 outperforms PAIR in the metrics of MRR@10 and Recall@5, we consider that dynamic listwise distillation enables the retriever to capture the re-ranker ability of  passage ranking at top ranks. Our model is trained with complete in-domain training data. Different from the baselines, we adopt a listwise training approach to jointly train both retriever and re-ranker and couple the two models by dynamic listwise distillation with hybrid data augmentation.

(2) We notice that different PLMs are used in different approaches, as shown in the second column of Table~\ref{tab:main_results}. In our approach, we use ERNIE base as the backbone model. We replacing BERT base used in DPR with ERNIE base to examine the effect of the backbone model, namely DPR-E. we observe that although both two methods employ the same backbone PLM, our method significantly outperforms DPR-E, indicating that PLM is not the factor for improvement.

(3) Among sparse retrievers, we find that COIL outperforms other methods, which seems to be a robust sparse baseline that gives substantial performance on the two datasets. 
We also observed that sparse retrievers overall perform worse than dense retrievers, such a finding has also been reported in prior studies~\cite{ANCE,MEBERT,rocketqa}, which indicates the effectiveness of the dense retrieval approach.

\begin{table*}[]
    \small
    \centering
    \begin{tabular}{lcccc}
    \toprule
    \textbf{Methods} & \textbf{PLM} & \textbf{\#candidate} & \textbf{Retriever} & \textbf{MRR@10}\\
    \midrule
    \tabincell{c}{BM25~(anserini)~\citep{bm25}}                & - & - & - & 18.7  \\
    \tabincell{c}{ColBERT~\citep{colbert2020sigir}} & BERT$_\text{base}$ & 1000 & BM25 & 34.9  \\
    \tabincell{c}{BERT$_\text{large}$~\citep{nogueira2019passage}} & BERT$_\text{large}$ & 1000 & BM25 & 36.5 \\
    \tabincell{c}{RepBERT~\citep{zhan2020repbert}}  & BERT$_\text{large}$ & 1000 & RepBERT & 37.7 \\ 
    \tabincell{c}{Multi-stage~\citep{multistage}}   & BERT$_\text{base}$ & 1000 & BM25 & 39.0 \\
    {CAKD~\citep{hostatter2020improving}}           & DistilBERT & 1000 & BM25 & 39.0 \\
    \tabincell{c}{ME-BERT~\citep{MEBERT}}           & BERT$_\text{large}$ & 1000 & ME-BERT & 39.5  \\ 
    \tabincell{c}{ME-HYBIRD~\citep{MEBERT}}         & BERT$_\text{large}$ & 1000 & ME-HYBIRD & 39.4  \\
    \tabincell{c}{TFR-BERT~\citep{Han2020learning}} & BERT$_\text{large}$ & 1000  & BM25 & {40.5}  \\
    RocketQA~\citep{rocketqa}        & ERNIE$_\text{base}$ & 50  & RocketQA & {40.9}  \\
    \midrule
    \multirow{3}*{\textbf{RocketQAv2 (re-ranker)}} & {ERNIE$_\text{base}$} & {1000}  & BM25 & 40.1 \\
    &{ERNIE$_\text{base}$} & {50}&RocketQA & {41.8} \\
    &{ERNIE$_\text{base}$} & {50}&RocketQAv2~(retriever) & \textbf{41.9} \\
    \bottomrule
    \end{tabular}
    \caption{The MRR@10 results of different methods for passage re-ranking on MSMARCO dataset. We copy the baseline results from original papers and report the PLM, candidate number and retriever for each method.}
    \label{tab:rerank}
\end{table*}

\subsection{Results on Passage Re-ranking}
In this part, we first describe the comparing baselines, then report the results on passage re-ranking.

\subsubsection{Baselines}
We report the results of the following baselines: BM25~\citep{bm25}, ColBERT~\citep{colbert2020sigir}, BERT$_\text{large}$~\citep{nogueira2019passage}, RepBERT~\citep{zhan2020repbert}, Multi-stage~\citep{multistage}, CAKD~\citep{hostatter2020improving}, ME-BERT~\citep{MEBERT}, ME-HYBIRD~\citep{MEBERT}, TFR-BERT~\citep{Han2020learning} and RocketQA~\citep{rocketqa}.
Among these methods, BM25 is a term-based method, and the rest are BERT-based methods based on neural networks. Since RocketQA does not report re-ranking results, we use the open-source re-ranker in RocketQA repository for evaluation. We report the results of RocketQAv2 re-ranker based on BM25 retriever with 1000 candidates, RocketQA retriever with 50 candidates and RocketQAv2 retriever with 50 candidates for comparing.

The prior works follow the two-stage approach (i.e., retrieve-then-rerank), where a passage retriever retrieves a (usually large) list of candidates from the passage collection in the first stage. In the second stage, a more expensive model (e.g., BERT-based cross-encoder) re-ranks the candidates. Note that the retrievers in baseline models may be differently designed.

\subsubsection{Results}
Table~\ref{tab:rerank} summarizes the passage re-ranking performance of RocketQAv2 re-ranker and all baselines on MSMARCO dataset.

As we can see, the RocketQAv2 re-ranker significantly outperforms all the competitive methods, demonstrating that the re-ranker benefits from our joint learning process, which is optimized to fit the relevance distribution of the retriever with dynamic listwise distillation.
Morever, if we use RocketQAv2 re-ranker to replace RocketQA re-ranker and apply it on the retrieval results by RocketQA retriever, we can see that RocketQAv2 re-ranker brings 0.9 percentage point improvement comparing to RocketQA re-ranker. This also demonstrates the effectiveness of RocketQAv2 re-ranker. 
Additionally, if we apply RocketQAv2 re-ranker on the top 1000 candidates by BM25, the performance is significantly better than other base models, and comparable to other large models. 


\begin{table}[]
    \small
    \centering
    \begin{tabular}{cccc}
    \toprule
    \textbf{Methods} & \textbf{MRR@10} & \textbf{R@50} \\
    \midrule
    \tabincell{c}{RocketQAv2~(retriever)} & \textbf{37.4}  & \textbf{84.9} \\
    \midrule
    \tabincell{c}{w/ Static Distillation}       & 36.0 & 84.5  \\
    \tabincell{c}{w/ Pointwise}       & 36.3 & 83.9  \\
    \tabincell{c}{w/o Denoised Instances}       & 36.3 & \textbf{84.9} \\
    \bottomrule
    \end{tabular}
    \caption{The results of different variants of RocketQAv2 retriever with eight training instances per query on MSMARCO dataset. Note that the results on NQ are similar and omitted here due to limited space.}
    \label{tab:ablation_study}
\end{table}

\subsection{Detailed Analysis}
Apart from the above illustration, we also implement detailed analysis on both dynamic listwise distillation and hybrid data augmentation.

\subsubsection{Analysis on Distillation}
\label{section:analysis_distill}
In this section, we analyze the results of retriever by replacing the optimization form in dynamic listwise distillation.

\paratitle{Dynamic or Static?}\quad
To examine the effect of dynamic optimization in distillation, we utilize a well-trained cross-encoder based re-ranker as a teacher model to perform static distillation comparing with dynamic listwise distillation. During static distillation, the parameters of re-ranker model are not updated and the retriever captures the relevance knowledge from the re-ranker in a traditional knowledge distillation manner.
As shown in Table~\ref{tab:ablation_study}, training with static distillation brings a performance drop. It demonstrates that dynamic optimization of both retriever and re-ranker enables to share relevance distributions with each other and brings a significant performance improvement.

\paratitle{Listwise or Pointwise?}\quad
To study the effect of the listwise training approach, we replace it with the pointwise training approach for the re-ranker during joint training. In such case, the training approaches of the retriever and the re-ranker are actually different. The re-ranker mainly optimized by the pointwise relevance scores of instances in $\mathcal{P}_q$, while the retriever has to learn the relevance ~\footnote{To enable learning the retriever, the pointwise relevance score from the re-ranker should be normalized in a listwise way. } from the re-ranker in a listwise way. 
Table~\ref{tab:ablation_study} shows that the pointwise training approach brings performance drop, and the listwise training approach performs better in our joint training archtecture. It demonstrates that the listwise training approach is more suitable in our joint training architecture than pointwise, since it can better simulate the relevance distribution in dynamic listwise distillation.

\subsubsection{Analysis on Hybrid Data Augmentation}
In this section, we conduct a detailed analysis for the hybrid data augmentation.

\paratitle{The Effect of Denoised Instances}\quad
In order to examine the effect of hybrid training data, we remove the denoised instances in training data and only use the undenoised data for joint training. Table~\ref{tab:ablation_study} shows the performance drop in terms of MRR@10 without denoised instances, which indicates that training data generated from different ways better represent the distribution of all the passages in the whole collection, and improve the performance especially on the metrics at top ranks.

\begin{figure}
	\centering
	\includegraphics[width=0.48\textwidth]{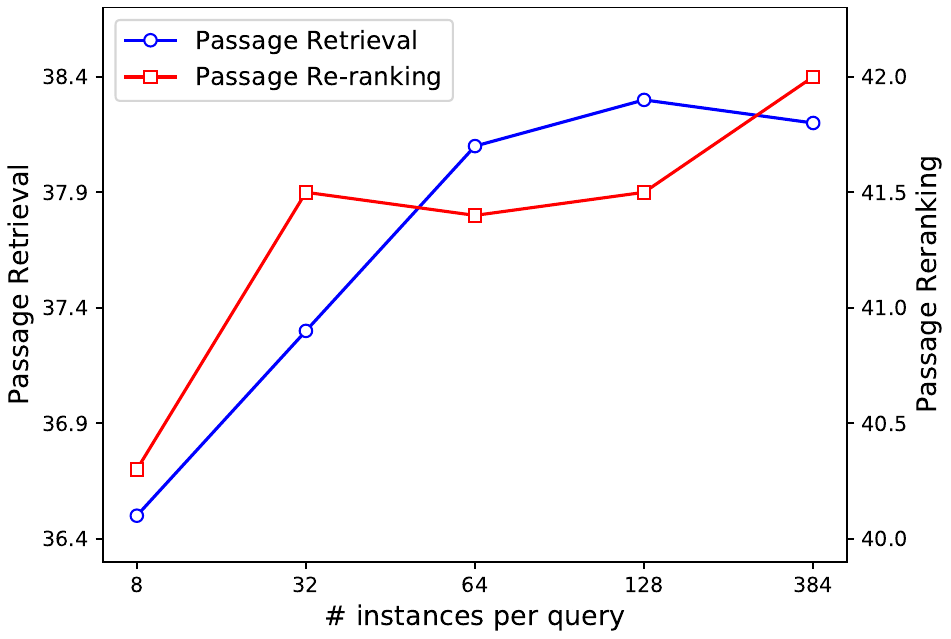}
	\caption{MRR@10 results of passage retrieval and passage re-ranking with different numbers of instances per query on MSMARCO. Note that instances per query contain one positive instance, and the rest are hard negatives.}
	\label{fig:negnum}
\end{figure}

\paratitle{The Number of Hard Negatives}\quad
 In hybrid data augmentation, we focus on obtaining diverse hard negatives. In our experiments, we find that the number of hard negatives significantly affects the performance of our joint training approach. As we described in previous section, for each query, we sample one positive instance and the rest of instances in the instance list $\mathcal{P}_q$ are hard negatives. Thus, the effect of the number of hard negatives should be equivalent to the effect of the number of instances. Figure~\ref{fig:negnum} shows the effect of the number of instances on both the passage retrieval and the passage re-ranking. From Figure~\ref{fig:negnum}, we can observe that a larger number of instances~(\ie number of hard negatives) improves the performance. The result demonstrates that instance list $\mathcal{P}_q$ with more instances can better represent the distribution of all the passages in the whole collection.


\paratitle{Incorporation of In-batch Negatives}\quad
For further study, we examine the effect of in-batch negatives in joint training process. Besides the hard negatives, we incorporate in-batch sampling during the joint training process, which can increase the amount of negatives for each query. Although the queries have additional in-batch negatives, we did not observe the performance improvements. 


\section{Conclusion}
This paper has presented a novel joint training approach for dense passage retrieval and passage re-ranking. 
To implement the joint training, we have made two important technical contributions, namely dynamic listwise distillation and hybrid data augmentation. 
Such an approach is able to enhance the mutual improvement between the retriever and the re-ranker, which can also simplify the training process.  
 Extensive results have demonstrated the effectiveness of our approach. To our knowledge, it is the first time that 
 the retriever and re-ranker are jointly trained in a unified architecture, which provides the possibility of  training the entire retrieval architecture in an end-to-end way.


\section*{Acknowledgements}
This work is partially supported by the National Key Research and Development Project of China (No.2018AAA0101900), National Natural Science Foundation of China under Grant No. 61872369, Beijing Outstanding Young Scientist Program under Grant No. BJJWZYJH012019100020098 and Public Computing Cloud, Renmin University of China.

\bibliography{anthology,ref}
\bibliographystyle{acl_natbib}




\end{document}